# A Multiple Kernel Learning Approach for Human Behavioral Task Classification using STN-LFP Signal *

Hosein M. Golshan, Adam O. Hebb, Sara J. Hanrahan,
Joshua Nedrud, Mohammad H. Mahoor, *Senior Member, IEEE*

*Abstract*— Deep Brain Stimulation (DBS) has gained increasing attention as an effective method to mitigate Parkinson's disease (PD) disorders. Existing DBS systems are open-loop such that the system parameters are not adjusted automatically based on patient's behavior. Classification of human behavior is an important step in the design of the next generation of DBS systems that are closed-loop. This paper presents a classification approach to recognize such behavioral tasks using the subthalamic nucleus (STN) Local Field Potential (LFP) signals. In our approach, we use the time-frequency representation (spectrogram) of the raw LFP signals recorded from left and right STNs as the feature vectors. Then these features are combined together via Support Vector Machines (SVM) with Multiple Kernel Learning (MKL) formulation. The MKL-based classification method is utilized to classify different tasks: button press, mouth movement, speech, and arm movement. Our experiments show that the $l_p$-norm MKL significantly outperforms single kernel SVM-based classifiers in classifying behavioral tasks of five subjects even using signals acquired with a low sampling rate of 10 Hz. This leads to a lower computational cost.

I. INTRODUCTION

Parkinson's disease (PD) is a chronic and progressive neurodegenerative disorder pertaining to the central nervous system [1]. Although, the main cause of this phenomenon is still unknown, some studies show the interaction of distinct processing circuits of the basal ganglia and cortex may be involved [1,2]. The symptoms of PD appear by the malfunction and death of dopamine-generating cells in an area of the brain called substantia nigra. The lack of these vital neurons causes various motor disorders including tremor, rigidity, bradykinesia, and postural instability [3].

Although, there is currently no certain cure for PD, there are different kinds of treatment options such as medication and surgery to alleviate the disorder manifestations. In recent years, Deep Brain Stimulation (DBS) has been considered as an effective treatment to deal with PD, specifically when drug therapy is no longer sufficient [1]. Using high frequency (~130-185 Hz) electrical pulses, DBS stimulates specific targets in the brain including the subthalamic nucleus (STN). This procedure is done through surgically implanted electrodes that are supplied by a battery-powered implanted pulse generator (IPG) [4].

Despite its remarkable performance in providing relief of PD's motor symptoms, DBS may cause side effects such as cognitive and balance disruptions [1]. This is mainly because of the existing open-loop DBS systems and lack of understanding about the mechanism of its action. In other words, altered physiological dynamics from adjusting the stimulation parameters, e.g., voltage, pulse duration and frequency, remains uncertain [5].

Hence, developing a closed-loop DBS system to be able to automatically adjust the stimulation parameters is currently an important research area. Contrary to the existing open-loop systems which provide a time invariant stimulation pulse, a closed-loop DBS system would generate a customized stimulation based on the patients' current behavior, reducing the undesirable side effects of the DBS therapy [1,5].

To design an efficient closed-loop DBS system, recognition of different human behavioral tasks based on feedback of brain signals is a main issue to address. Note that, DBS surgery provides a suitable opportunity to gain access to different brain signals including the Local Field Potential (LFP) signal, and record them directly from basal ganglia [1,6]. Different human activities are coded in the LFP signal, so these activities may be characterized by processing the collected signal.

So far, there have been several studies involving detection/classification of different human behavioral tasks based on different kinds of brain signals. Many algorithms have been developed based on the processing of electroencephalography (EEG) and electrocorticography (ECoG), e.g., P300 speller paradigm [7,8], seizure detection [9], brain-switch based on motor imaginary [10]. Neural Networks [11,12] and SVM-based classifiers together with wavelet-domain feature extraction have been successful in detecting human behavior [5,13,14]. Most recently, STN-LFPs have been considered as another useful neural feedback signal to recognize human behavioral activities. Time-frequency analysis of motor cortex recorded with ECoG and STN-LFPs has shown that PD exhibits oscillatory behavior modulated by motor activities, which results in suppression of $\beta$ (13-35 Hz) frequency spectral power during motor tasks [1]. Taking this point into account, an SVM-based classifier was proposed in [15]. A motor task detection method using LFPs and non-linear regression has been addressed in [14].

* This research is partially supported by the Knoebel Institute for Healthy Aging at University of Denver.

Hosein M. Golshan is a PhD student in the ECE department at the University of Denver, Denver, CO. (email: hosein.golshanmojdehi @du.edu)

Adam O. Hebb is a neurosurgeon at Colorado brain and spine institute, Englewood, CO, and a research scholar in the ECE department, University of Denver, Denver, CO.(email: adam.hebb@aoh.md)

Sara J. Hanrahan is an investigator scientist at Colorado Neurological Institute (CNI), Englewood, CO.(email: shanrahan@thecni.org)

Joshua Nedrud is an investigator scientist at Colorado Neurological Institute (CNI), Englewood, CO.(email: jnedrud@thecni.org)

Mohammad H. Mahoor is an Associate Professor in the ECE department at the University of Denver, Denver, CO. (email: mohammad.mahoor@du.edu)

An adaptive learning approach based on LFP signal was suggested in [16], where a hybrid model for combining SVM and hidden Markov model was used for human behavioral clustering.

In this paper, we present an $l_p$-norm Multiple Kernel Learning (MKL) [17] approach for recognition of different human behavioral activities using STN-LFPs collected from five subjects undergoing DBS surgery. The spectrogram of the raw LFPs related to each event is used to extract the feature vectors. In contrast to the previously presented single kernel SVM-based classifiers, the proposed MKL approach utilizes the corresponding LFPs from right and left STNs simultaneously. This in return provides a higher performance with the proposed method. Our experimental results show that the classification accuracy of the presented scheme is robust even if the feature vectors are largely down-sampled. So, this leads to a lower computational burden.

The rest of this paper is organized as follows: Section II illustrates the data recording procedure. Section III presents the $l_p$-norm MKL method as well as the proposed approach. Section IV gives the comparisons and quantitative results. Finally, conclusions and remarks are discussed in Section V.

## II. DATA RECORDING

Five subjects undergoing DBS surgery participated in this study. All the participants provided informed consent as approved by the HealthOne Institutional Review Board. Five bilateral recordings were performed using DBS leads implanted in both STNs. In addition, we did not proceed with recording until patients were fully awake during surgery. Note that, all patients were in the off medication state [1].

LFP signals were recorded from all four contacts of each DBS lead (*Medtronic, MN, USA*). The collected data were then amplified, digitized (5 kHz), band passed filtered (1-100 Hz), and combined with event markers and subject responses. Meanwhile, a linked-mastoid common reference was used for recordings. Finally, the LFP channels were bipolar re-referenced (0-1, 1-2, 2-3) before any analysis [1].

The experiments included four different tasks: button press, mouth movement, speech, and arm movement. For each task, a block of several repetitions was designed. In terms of the task initiation, subjects were cued by an audio signal. For the "button press", the patients were asked to press a button using either the left or right thumb. "Speech" task consisted of reciting simple object names displayed on screen. For "arm movement" task, the patients had to raise their arm to reach a highlighted target. Finally, "mouth movement" simply comprised moving the mouth without speech as a comparison to speech trials.

## III. METHODOLOGY

In this section, a brief review of the $l_p$-norm MKL classifier is first given. Then, the feature extraction approach as well as the proposed classification scheme is presented.

### A. $l_p$-norm Multiple Kernel Learning

The MKL method has gained much popularity in the pattern recognition and machine learning community due to its desirable classification ability. Recent works [17,18] show that MKL can improve the discriminant power of the SVM classifier. The main idea behind MKL is to optimally combine matrices calculated based on multiple features with multiple kernels in SVM [17]. Substantially, kernel functions map the features to a new space where they can linearly be separable. The MKL-based SVM aims to learn both the decision boundaries between different classes and kernel combination weights in a single optimization problem [18].

Here, we employ a recently presented realization of the MKL classifier called $l_p$-norm MKL, which proved to be more flexible in selecting different kernel combinations. An $l_p$-norm MKL ($p \geq 1$) is defined as follows [17]:

$$\min_{\omega,\omega_0,\xi} (\omega,\omega_0,\xi) = \frac{1}{2}\|\omega\|_{2,p}^2 + C\sum_{i=1}^{N}\xi_i \qquad \xi_i \geq 0, p \geq 1$$

$$\text{s.t. } y_i\left(\sum_{m=1}^{M}\omega_m^T \phi_m(x_i) + \omega_0\right) \geq 1-\xi_i, \qquad i=1,2,...,N \quad (1)$$

where, $\phi_m(\cdot)$ maps the feature vector $x_i$ to another space based on which the kernel function $k(\cdot,\cdot) = \langle\phi_m(\cdot),\phi_m(\cdot)\rangle$ is defined. $\{\omega_m\}$s are the parameters of the decision hyper-planes. $M$ and $N$ are the number of kernels and training samples respectively. $C$ is the penalty parameter and $\xi_i$ is the slack parameter.

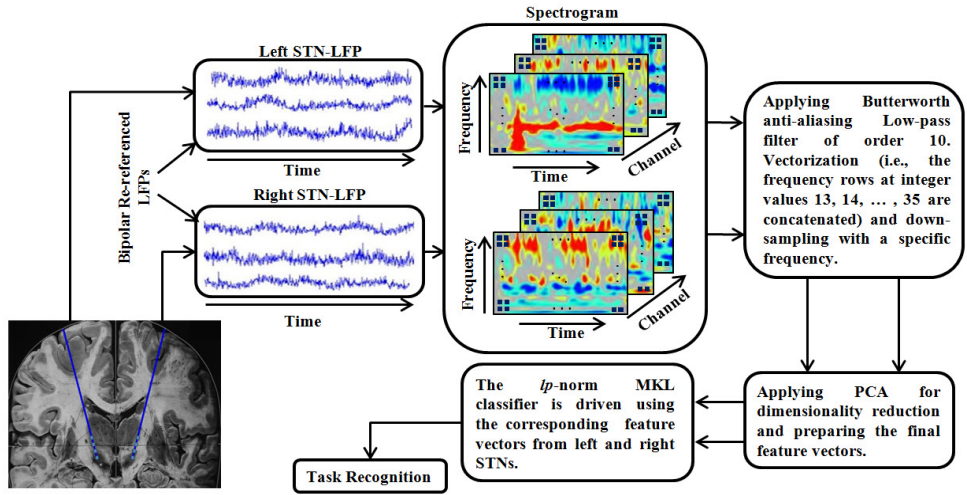

Figure. 1. Representation of the proposed human behavioral task classification based on STN-LFPs. Black squares on each spectrogram show how the time-frequency coefficients of the corresponding bipolar signal are down-sampled.

This convex optimization problem is solved using its dual form as follows [17]:

$$\min_d \max_\alpha L(d,\alpha) = 1^T\alpha - \frac{1}{2}\alpha^T Y K_d Y \alpha, \quad p \in [1,2)$$

$$\max_d \min_\alpha L(d,\alpha) = 1^T\alpha - \frac{1}{2}\alpha^T Y K_d Y \alpha, \quad p \in (2,+\infty) \quad (2)$$

$$\text{s.t. } K_d = \sum_{m=1}^M d_m K^{(m)}, \sum_{i=1}^N \alpha_i y_i = 0, 0 \leq \alpha_i \leq C, \sum_{m=1}^M d_m^{p/(2-p)} \leq 1, d_m \geq 0$$

where, $\alpha = (\alpha_1, ..., \alpha_N)^T$ is the vector of Lagrangian dual variables, $Y = \text{diag}(y_1, ..., y_N)$ is an $N \times N$ diagonal matrix ($y_i$ is the label of each sample $x_i$), $K^{(m)}$ is the kernel matrix corresponding to the $m^{th}$ kernel function. $d = (d_1, ..., d_N)^T$ is the kernel combination vector that controls the weight of $(\|\omega_m\|^2)$ in the objective function of Eq. (1). The other parameters are defined exactly the same as given for Eq. (1).

### B. Feature Extraction

As mentioned earlier, here, we use the time-frequency representation (spectrogram) of the collected STN-LFPs as our features. It has been shown that different behavioral tasks yield different representations in the time-frequency domain [1]. Thus, it is an appropriate measure to differentiate various human behaviors. To obtain the spectrogram of a continuous signal, we apply the continuous wavelet transform (CWT) on the bilaterally re-referenced LFPs. We employ the complex Morlet (C-Morlet) mother wavelet which is proven to be a suitable choice for biomedical signal processing [1,15]:

$$X_\omega(a,b) = \int_{-\infty}^{+\infty} \frac{x(t)}{\sqrt{a}} \psi(\frac{t-b}{a}) dt, \quad \psi(t) = \frac{e^{-t^2/f_b}}{\sqrt{\pi f_b}} e^{j2\pi f_c t} \quad (3)$$

where, $X_\omega(a,b)$ is the CWT of the function $x(t)$ with two variables $a$ (scaling parameter) and $b$ (shift parameter). $\psi$ is the C-Morlet mother wavelet. $f_c$ and $f_b$ are respectively the wavelet center frequency and bandwidth parameter. Since we are interested in analyzing the $\beta$ frequency components, $f_c$ is set in the range of (13-35 Hz) here.

### C. Classification Scheme

The first step to acquire the training and test samples is to specify those parts of the raw LFP data that are related to different events. A time window ranging from 1 sec before to 1 sec after each onset is used to determine the relevant time interval. Afterwards, the corresponding wavelet coefficients in the $\beta$ frequency range are calculated using Eq. (3).

To drive the $l_p$-norm MKL classifier, we consider the right and left spectrograms for each event as two feature vectors. First, the two dimensional spectrograms are converted to vectors, and low-pass filtered by an anti-aliasing Butterworth filter of order 10. Then, these feature vectors are down-sampled to keep the computational cost low. Finally, the MKL classifier is applied on these pre-processed data to recognize the behavioral task related to each event. Figure. 1 shows the aforementioned procedure graphically.

## IV. EXPERIMENTAL RESULTS

To evaluate the accuracy of the presented approach, we use the raw LFP data from five different subjects, as described in Section II. The quantitative results of our method are compared with a recently proposed single kernel SVM-based method [15]. We also assess the effect of three different kernel functions on the SVM classifier separately: 1. Radial Basis Function (RBF) $k(x, y) = exp(\gamma\|x-y\|^2)$, 2. Linear function $k(x, y) = x^T y + c$, and 3. Polynomial function $k(x, y) = (x^T y + c)^d$. Note that, x and y are two feature vectors, and $\gamma$, c, and d are optional constants. In terms of the $l_p$-norm MKL, we set the parameter $p=1.8$, and for the SVM-based classifier all the parameters are set so as to achieve the best performance. A 10-fold cross validation is implemented in all experiments. Furthermore, in all cases, principal component analysis (PCA) is applied to reduce the dimensionality of data (in each case, 95% of the eigen-values corresponding to the maximum variance direction is kept).

Figure. 2 depicts the classification accuracy of different classifiers vs the sampling frequency (from 5kHz to 2Hz). As seen, the presented $l_p$-norm MKL classifier outperforms the other compared classifiers. In the experiments, we observed that the results of the MKL classifier are robust even when the sampling frequency of the feature vectors is drastically low. To measure the robustness of different methods against the size of the feature vectors, all the experiments are redone for different down-sampling rates. Table I gives the average classification accuracy of all five subjects for different scenarios (i.e., three and five-task classification and different sampling frequencies). As shown, regardless of the sampling frequency and the number of tasks, the presented MKL-based classifier returns the best results. Figure. 3 provides the average confusion matrix of all subjects with sampling frequency of 10Hz and five-task classification, which summarizes the identification results.

TABLE I. COMPARISON OF CLASSIFICATION ACCURACY (%). THE BOLD VALUE IS THE BEST ONE IN EACH CASE. 3 TASKS: BUTTON PRESS, SPEECH, AND REANDOM SEGMENT. 5 TASKS: BUTTON PRESS, ARM MOVEMENT, SPEECH, MOUTH MOVEMENT, AND RANDOM SEGMENT.

| Sampling Frequency | 3 Tasks | | 5 Tasks | |
|---|---|---|---|---|
| | SVM-Linear | MKL | SVM-Linear | MKL |
| 5000 (Hz) | 66.92 | **68.75** | 53.17 | **54.86** |
| 500 (Hz) | 66.40 | **68.66** | 53.43 | **54.19** |
| 50 (Hz) | 65.44 | **70.39** | 53.33 | **56.56** |
| 25 (Hz) | 65.45 | **70.31** | 53.27 | **57.44** |
| 10 (Hz) | 66.40 | **70.83** | 51.29 | **58.34** |
| 2 (Hz) | 58.15 | **65.33** | 44.05 | **53.53** |

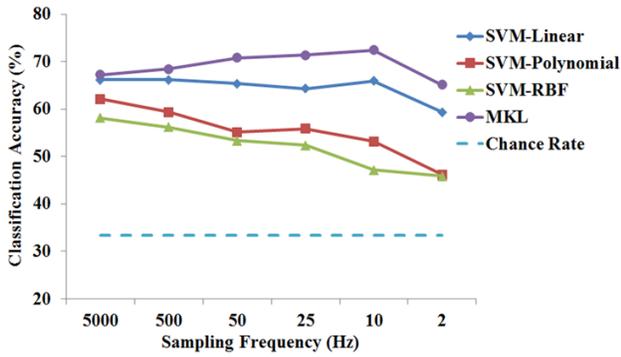
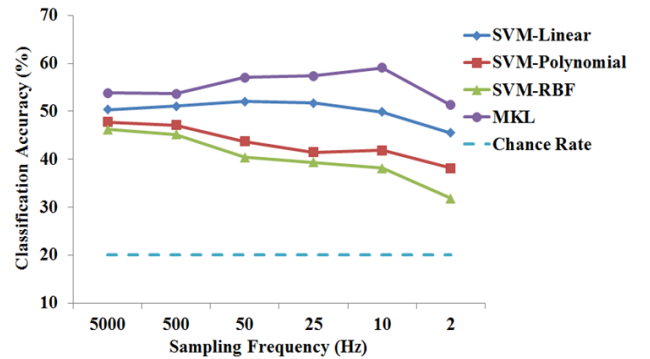

Figure. 2. The classification accuracy (%) of different methods. Left and right graphs respectively show the results for 3-task (Speech, Button press, Random segment) and 5-task (Speech, Button press, Arm movement, Mouth movement, Random segment) classification. The reason behind using random segments is to train the classifier to recognize other tasks rather than the existing ones. The "Chance Rate" is to show the qualification of each classifier. If the accuracy is below the "Chance Rate", it means that the classifier is not a suitable choice; it is nothing but a random operator.

In terms of execution time of the algorithms, using MATLAB 2013a (*Mathworks Inc*) on a PC with Intel Ci5 CPU (3.4GHz) and 8GB memory, the average training time of each fold of the MKL method for the sampling frequency of 5kHz and 10Hz are about 450s and 6s respectively. The execution time in the test phase for each sample is about 1.5ms and 1ms respectively for the 5kHz and 10Hz sampling rates. For the single kernel SVM classifier, however, the average training time of each fold is about 50s and 0.2s respectively for the aforementioned sampling frequencies, and testing each sample takes about 2ms and 0.3ms. Note that, while 6GB RAM is needed for the 5kHz sampling rate, only 0.25GB memory is used for the 10Hz rate.

## V. CONCLUSION AND DISCUSSION

In this paper, an $l_p$-norm MKL approach for classification of different human behavioral tasks using STN-LFP signal was presented. A feature extraction method based on the time-frequency analysis (spectrogram) of the collected signal was developed. We used the left and right LFPs acquired from the corresponding STNs as two feature vectors. This led to a higher performance with the MKL classifier. The experiments were conducted on five dataset recorded from patients undergoing DBS surgery. The quantitative results confirmed the superiority of the proposed method in almost all cases.

In addition, to evaluate the robustness of the classifiers with respect to the size of feature vectors, different down-sampling rates were tested. In contrast to the single kernel SVM-based methods, the proposed MKL approach shows promising results even for very low sampling frequencies. This led to a lower computational burden. Note that, a more accurate classification of the human behavioral tasks would be a precursor for developing future closed-loop DBS systems, which is a cutting-edge research area.

Developing a more robust feature extraction method is an interesting extension to this work. Moreover, evaluating the connectivity of different bipolar channels across the DBS leads can be another relevant topic for future research.

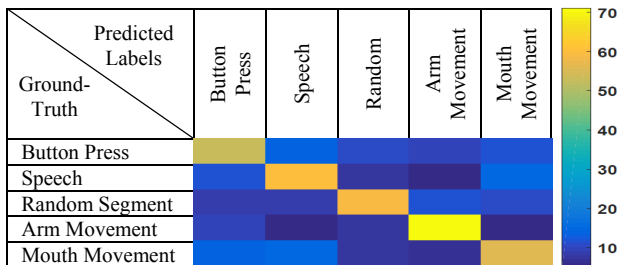

Figure 3. Average confusion matrix of five subjects. The results are reported for the MKL classifier and sampling frequency of 10 Hz. All the values are normalized in the range of [0 100].